\documentclass[review]{elsarticle}
\graphicspath{ {./figures/} }
\usepackage{hyperref}
\usepackage{float}
\usepackage{verbatim} 
\usepackage{apalike}
\restylefloat{figure}
\usepackage{ragged2e}

\graphicspath{ {./figures/} }
\usepackage[table]{xcolor}
\usepackage{verbatim} 
\usepackage{apalike}
\usepackage{caption}
\usepackage{subcaption}
\usepackage[margin=1in]{geometry} 
\usepackage[]{graphicx}
\usepackage[font=small,skip=0pt]{caption}
\usepackage{enumitem}
\usepackage{booktabs}
\usepackage{longtable}
\usepackage{tabularx}
\usepackage{cancel}
\usepackage{multirow}
\usepackage{supertabular}
\usepackage{algorithmic}
\usepackage{algorithm}
\usepackage{amsmath}
\usepackage{rotating}
\usepackage[normalem]{ulem}
\usepackage[utf8]{inputenc}
\usepackage[T1]{fontenc}
\usepackage{lineno}
\usepackage{array}
\usepackage{makecell}
\usepackage{pdflscape}
\usepackage{afterpage}
\usepackage{capt-of}
\usepackage{lipsum}
\usepackage{changepage}
\usepackage{xcolor}
\usepackage{rotating} 
\usepackage{graphicx}
\usepackage{adjustbox}

\usepackage{amssymb}
\usepackage{natbib}
\bibpunct{(}{)}{,}{a}{}{;}
\renewcommand{\citep}[1]{(\citealp{#1})}

\journal{}
\biboptions{authoryear}

\begin{document}
\begin{frontmatter}

\begin{titlepage}
\begin{center}
\vspace*{1cm}

\textbf{ \large Attention is all you need for an improved CNN-based flash flood susceptibility modeling. The case of the ungauged Rheraya watershed, Morocco}

\vspace{1.5cm}

Akram Elghouat$^{a}$ $^{b}$, Ahmed Algouti $^{a}$, Abdellah Algouti $^{a}$, Soukaina Baid $^{a}$

\hspace{10pt}

\begin{flushleft}
\small  
$^a$ Cadi Ayyad University, Faculty of Sciences Semlalia, Department of Geology, 2GRNT Laboratory (Geosciences, Geotourism, Natural Hazards and Remote Sensing). Marrakech, Morocco. 

$^b$ Interuniversity Institute for Earth System Research (IISTA), University of Granada. 18006, Granada, Spain. 
 \\

\vspace{1cm}

elghouatakram@go.ugr.es , algouti@uca.ac.ma, abalgouti@uca.ac.ma, soukaina.baid@ced.uca.ma

\vspace{1cm}
\textbf{Corresponding Author: elghouatakram@go.ugr.es (A. Elghouat)} \\

\end{flushleft}        
\end{center}
\end{titlepage}



%



\begin{abstract}

Effective flood hazard management requires evaluating and predicting flash flood susceptibility. Convolutional neural networks (CNNs) are commonly used for this task but face issues like gradient explosion and overfitting. This study explores the use of an attention mechanism, specifically the convolutional block attention module (CBAM), to enhance CNN models for flash flood susceptibility in the ungauged Rheraya watershed, a flood prone region. We used ResNet18, DenseNet121, and Xception as backbone architectures, integrating CBAM at different locations. Our dataset included 16 conditioning factors and 522 flash flood inventory points. Performance was evaluated using accuracy, precision, recall, F1-score, and the area under the curve (AUC) of the receiver operating characteristic (ROC). Results showed that CBAM significantly improved model performance, with DenseNet121 incorporating CBAM in each convolutional block achieving the best results (accuracy = 0.95, AUC = 0.98). Distance to river and drainage density were identified as key factors. These findings demonstrate the effectiveness of the attention mechanism in improving flash flood susceptibility modeling and offer valuable insights for disaster management.\end{abstract}


\end{frontmatter}

\section{Introduction}

Flash floods are considered one of the most devastating, costly, challenging, and widespread natural hazards globally (\citealp{jeyaseelan2003},\citealp{tingsanchali2012urban}; \citealp {kuenzer2013flood}; \citealp{dahri2017monte}; \citealp{nogueira2018exploiting}; \citealp{mohanty2020flood}). These events emerge from rapid and excessive rainfall occurring within a concise timeframe, typically around 6 hours, and often in mountainous environments \citep{li2019risk}, resulting in erosion, landslides, substantial damage to infrastructure and properties, and even significant loss of life \citep{cao2016flash}. In fact, they are responsible for almost 84\% of global deaths related to natural disaster\citep{jamali2020rainwater}. In recent years, scientific communities worldwide have expressed huge concern about flash floods, as they expect that global warming, population growth, deforestation, and land use planning will increase the frequency and severity of these events, which will result in significant economic and social impacts (\citealp{bouwer2010changes}, \citealp{de2021ecosystem}, \citealp{mackay2008climate}, \citealp{woodruff2013coastal}, \citealp{hirabayashi2013global}, \citealp{guha2016dat}, \citealp{mekonnen2016four}).Therefore, there is an urgent need for effective spatial prediction of flash flood occurrences on different spatial scales to mitigate their damaging impacts.

Achieving absolute prevention of flood damage seems like a challenging and complex task. Flood management efforts therefore increasingly prioritize proactive measures like evaluating susceptibility and vulnerability in flood-prone regions \citep{hoque2019assessing}. The comprehensive understanding and identification of these susceptible areas through advanced and appropriate mapping techniques can offer effective mitigation strategies, including emergency evacuation plans (\citealp{guerriero2020flood}; \citealp{klipalo2022full}; \citealp{musolino2020comparison}; \citealp{tehrany2013spatial}; \citealp{cloke2009ensemble}). Flash flood susceptibility mapping (FFSM) is considered a crucial tool in preventing and managing flash flood events, as it provides significant information about the spatial likelihood of future flash floods taking into consideration topographical features, hydrological aspects, and climatic conditions (\citealp{jacinto2015continental}; \citealp{vojtek2019flood}). This provides targeted guidance to facilitate the implementation of flash flood control and disaster reduction measures that can subsequently reduce potential losses (\citealp{choubin2019ensemble}; \citealp{hong2018application}; \citealp{saha2021far}). Modeling susceptibility to flash floods is thus a prominent global research focus nowadays.

Multiple methods have been developed by researchers to assess and measure flood susceptibility (\citealp{sahana2020exploring};\citealp{costache2020comparative};\citealp{islam2021flood} \citealp{islam2021flood};\citealp{pradhan2023spatial}). Former approaches primarily relied on expert knowledge, physical-based models such as HEC-RAS, SWAT, WetSpa (\citealp{khalil2017floodplain}; \citealp{ongdas2020application}), and traditional statistical models, encompassing techniques like frequency ratio analysis, discriminant analysis, and generalized linear models (GLMs) (\citealp{choubin2019ensemble}; \citealp{zeng2021utilizing}. However, these approaches are time-consuming and require extensive measurement datasets (\citealp{costache2022flash}). Recently, the rapid evolution and advancements of technology in both hardware and software have opened the door to the use of more sophisticated and advanced data-driven models in environmental studies. Particularly, several machine learning (ML) algorithms have gained widespread application, such as artificial neural networks (ANN), support vector machines (SVM), random forests (RF), decision trees (DT), naive bayes (NB), and K-Nearest Neighbors (KNN) (\citealp{kabir2020deep};\citealp{pham2021flood};\citealp{bui2020verification} \citealp{wang2020flood};\citealp{eini2020hazard};\citealp{lin2022predicting};\citealp{khoirunisa2021gis}; \citealp{priscillia2021flood};\citealp{khosravi2018comparative}). These models have showcased superior performance and improved assessment quality compared to traditional methods (\citealp{elghouat2024integrated}; \citealp{khosravi2018comparative};\citealp{hayashi2019right};\citealp{gudiyangada2020novel}; \citealp{zhongping2020susceptibility}) . They analyze topography, hydrological traits, land use, and historical flood occurrences to identify key variables influencing flash floods. Moreover, the availability of remote sensing data has facilitated the application of ML techniques. These data sources provide vital information for mapping flood susceptibility by capturing details about past flood events in addition to different linked factors (\citealp{zhao2020urban}). This will help improve our understanding and prediction of floods, contributing to better flood management and mitigation strategies.

Deep learning methods, a subset of ML algorithms, have been increasingly explored for supervised and unsupervised classification, segmentation, and regression tasks. They prove particularly efficient and beneficial when dealing with large datasets (\citealp{wang2020flood};\citealp{wang2020comparative} ). Among DL models, deep neural networks (DNNs) and convolutional neural networks (CNNs) take precedence in applications due to their capacity and efficiency to automatically identify crucial features in datasets (\citealp{alzubaidi2021review}). For flood susceptibility, \cite{bentivoglio2022deep} highlighted that CNNs are the most utilized deep learning models across a majority of research papers. They showed superior predictive performance over conventional ML algorithms. For instance, \cite{zhao2020urban} conducted an urban flood susceptibility study utilizing RF, SVM, and CNN methods, with CNN demonstrating the highest performance. \cite{wang2020flood} compared SVM and CNN models for mapping flood susceptibility using different data-input methods, revealing CNN's superior accuracy in the 2D-based input approach. Similarly, \cite
{liu2021hybrid} assessed the performance of three hybrid models, including SVM, classification and regression trees (CART), and a CNN, with the latter outperforming the other models. These models consistently exhibit strong performance in flood susceptibility studies. Various CNN architectures, such as LeNet-5, AlexNet, VGGNet, ResNet, InceptionNet, and more, have been developed, each with distinct strengths and weaknesses. However, although these CNN architectures are robust and efficient, they still suffer from drawbacks (\citealt{joshi2019issues}). Deeper CNNs commonly face challenges such as gradient explosion or vanishing (\citealt{joshi2019issues}), whereas wider CNNs may induce overfitting phenomena (\citealt{joshi2019issues}). Therefore, the key is to ensure that CNNs concentrate on learning and emphasizing important information instead of learning non-useful background information.

In this context, attention modules have emerged as a significant enhancement for CNN architectures in multiple tasks, including image classification (\citealt{vaswani2017attention}). Along with attention modules, CBAM (i.e., convolutional block attention module) integrates both channel and spatial attention mechanisms within any CNN, allowing the network to focus not only on critical spatial features but also on important channels within the data (\citealt{woo2018cbam}). It is a lightweight module that adds very few extra parameters to a CNN and doesn't burden the computations much. Multiple studies have showcased the ability of this module to direct network focus towards learning only the important things, resulting in enhanced performance of their DL models (\citealt{zheng2022attention}; \citealt{ravi2023attention}; \citealt{mao2023attention}; \citealt{alirezazadeh2023improving}). However, the literature shows no studies with a focus on flash flood susceptibility modeling. Adding the attention mechanism could therefore make CNNs better at distinguishing and recognizing patterns in data related to flash floods, leading to more accurate results.

This study aims to assess the effectiveness of attention-based CNNs in predicting flash flood susceptibility. The specific objectives were: (1) to compare the performance of CBAM-based CNNs with baseline models such as ResNet18, DenseNet121, and Xception in predicting flash flood susceptibility and use the most accurate model to create a susceptibility map; and (2) to assess the impact of the CBAM attention module's location within the CNN architecture on flash flood susceptibility predictions. In this study, we selected the ungauged Rheraya watershed, a region with a history of flash flood events, and considered 16 well-chosen conditioning factors that influence flash flooding. The innovations of this study include: (1) it is the first application of attention-based CNNs in flash flood susceptibility; (2) it is the first evaluation of the CBAM attention module's location within the CNN architecture on flash flood susceptibility model performance; and (3) the generation of the first deep learning-based flash flood susceptibility map for the Rheraya watershed, a flood-prone area. The findings of this study will assist authorities and policymakers in mitigating the risks of flash flooding in the area and in devising effective measures to prevent potential losses.

\section{Materials and Methods} \label{Methodology}
\subsection{Study area} \label{Methodology}

\begin{figure}[H]
\centering
\includegraphics[width=15cm]{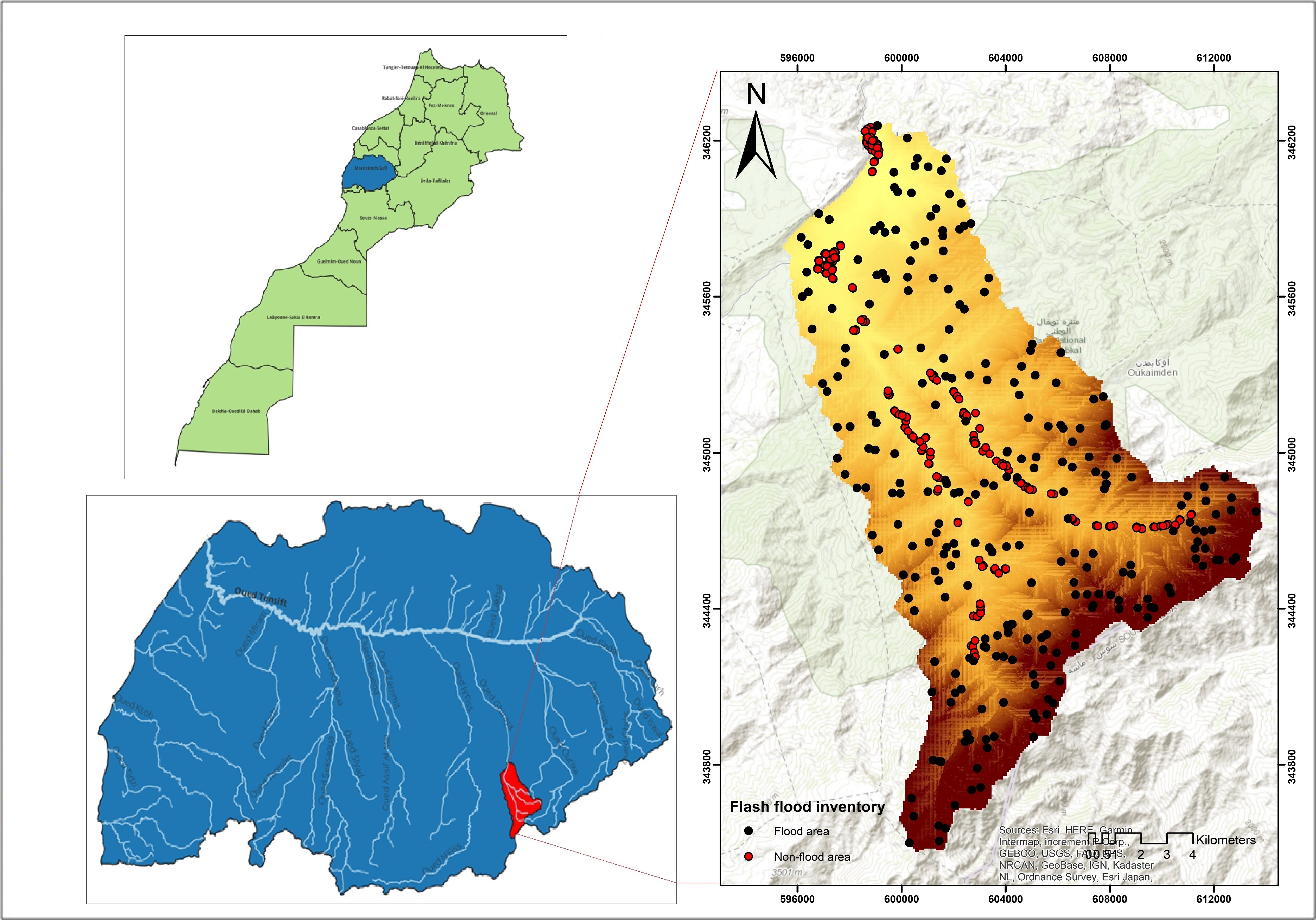}
\caption{Locations of the study area and historical flash floods}
\label{Figure1}
\end{figure}

The Rheraya watershed lies in the southern region of Marrakech city, within the Western High Atlas area. Its outlet is defined by the Tahanaout hydrometric station at coordinates 31.3°N, 7.9°E (Figure \ref{Figure1}). Key population centers in this region include Asni, Imlil, and Moulay Brahim. Regarded as a significant catchment area in the High Atlas range, the Rheraya covers approximately 224 km2, with elevations spanning from 1084 to 4167 meters and steep slopes reaching up to 80°. The region experiences an average annual rainfall of 363 mm, with temperatures ranging typically between 18 and 38 degrees Celsius. Hillslopes within this area are characterized by degraded rangelands with sparse vegetation and rocky terrain, while the valley hosts a narrow stretch of cultivated land on both sides of the river (\citealt{boudhar2007analyse}).

\begin{figure}[H]
\centering
\includegraphics[width=15cm]{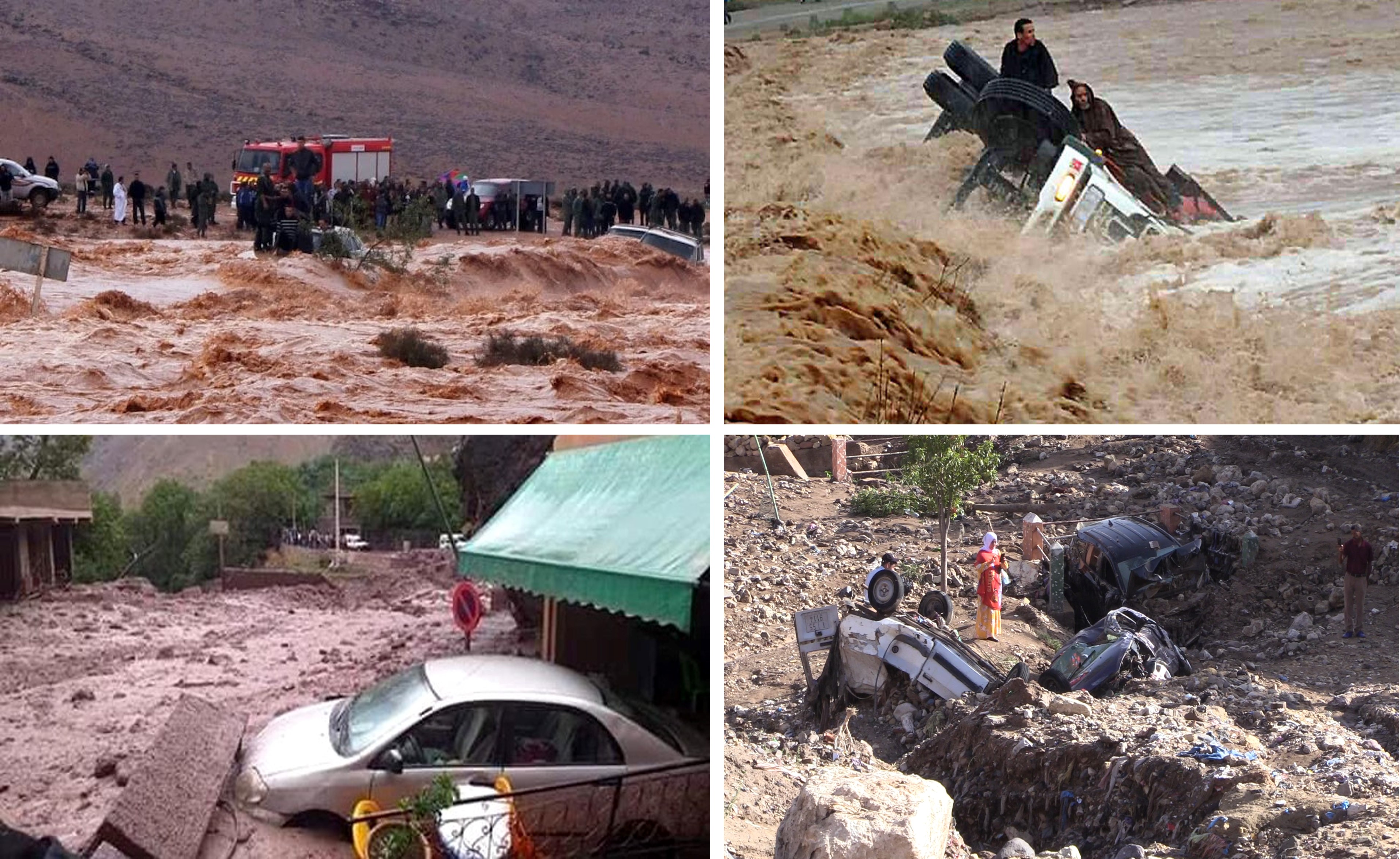}
\caption{Flash flood events and damages in the study area}
\label{Figure2}
\end{figure}

The study area is an ungauged basin with a documented history of destructive flash flood events causing substantial damage to agricultural land, infrastructure, and buildings. Notably, the region experienced a devastating flash flood in 1995 due to intense, short-term rainfall, resulting in over 150 fatalities, including 60 tourists (\citealt{digby2000changing}). Additionally, a severe flash flood in 2019 caused significant property damage (Figure \ref{Figure2}). The high vulnerability and loss of life in this area are linked to the population's occupation of these exposed regions, primarily for tourism purposes (\citealt{elfels2018flood}). The severity of the consequences of these events and the high frequency of this phenomenon in this region made it suitable for the implementation and development of an attention-based CNN model for flash flood susceptibility in order to evaluate its efficiency and predictive performance.

\subsection{Flash flood inventory} \label{Methodology}
Identifying and creating a precise historical inventory of flash floods is an essential step in assessing and investigating the relationship between flash floods and their causal factors, enabling the development of accurate flash flood susceptibility maps (\citealt{choubin2019ensemble}). Current approaches rely on historical archives, field surveys, and remote sensing satellite data. In this study, we generated an inventory of historical flash flood events based on records from the Hydraulic Basin Agency of Tensift (ABHT), followed by a field survey in which local residents contributed to the development of this inventory. Validating the obtained data is a mandatory step to ensure its trustworthiness and accuracy. Therefore, pre-flood and post-flood remote-sensing images were analyzed to ascertain the existence and validation of the documented flash flood records. We calculated the Sentinel-1 dual-polarization water index (SDWI) and the normalized difference water index (NDWI) using Landsat data to identify the flooded areas (\citealt{devries2020rapid}; \citealt{mehmood2021mapping}). The validation step was conducted using the Google Earth Engine platform (GEE). 

A total of 261 flash flood locations from 1994 to 2022 were identified. Additionally, we generated 261 points designated as 'non-flash flood' from areas where there was no evidence of flash flood occurrences using the geo-processing tool 'create random points', within the data management tool in ArcGIS software (\citealt{esri2011arcgis}).

\subsection{Flash conditioning factors} \label{Methodology}
\begin{figure}[H]
\centering
\includegraphics[width=11cm]{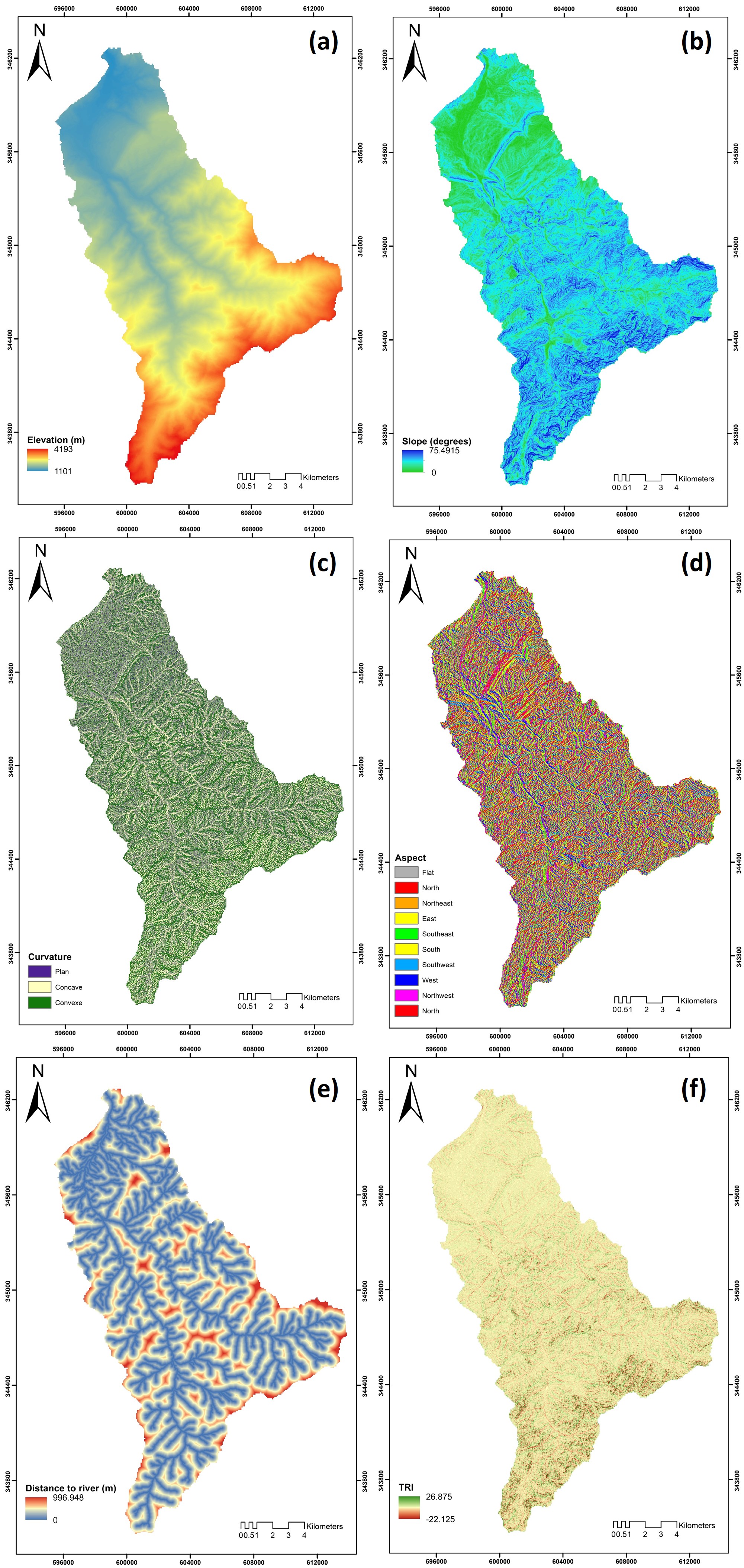}
\caption{(a) Elevation, (b) slope angle, (c) Curvature, (d) Aspect, (e) Distance to river, (f) TRI}
\label{Figure3}
\end{figure}

\begin{figure}[H]
\centering
\includegraphics[width=11cm]{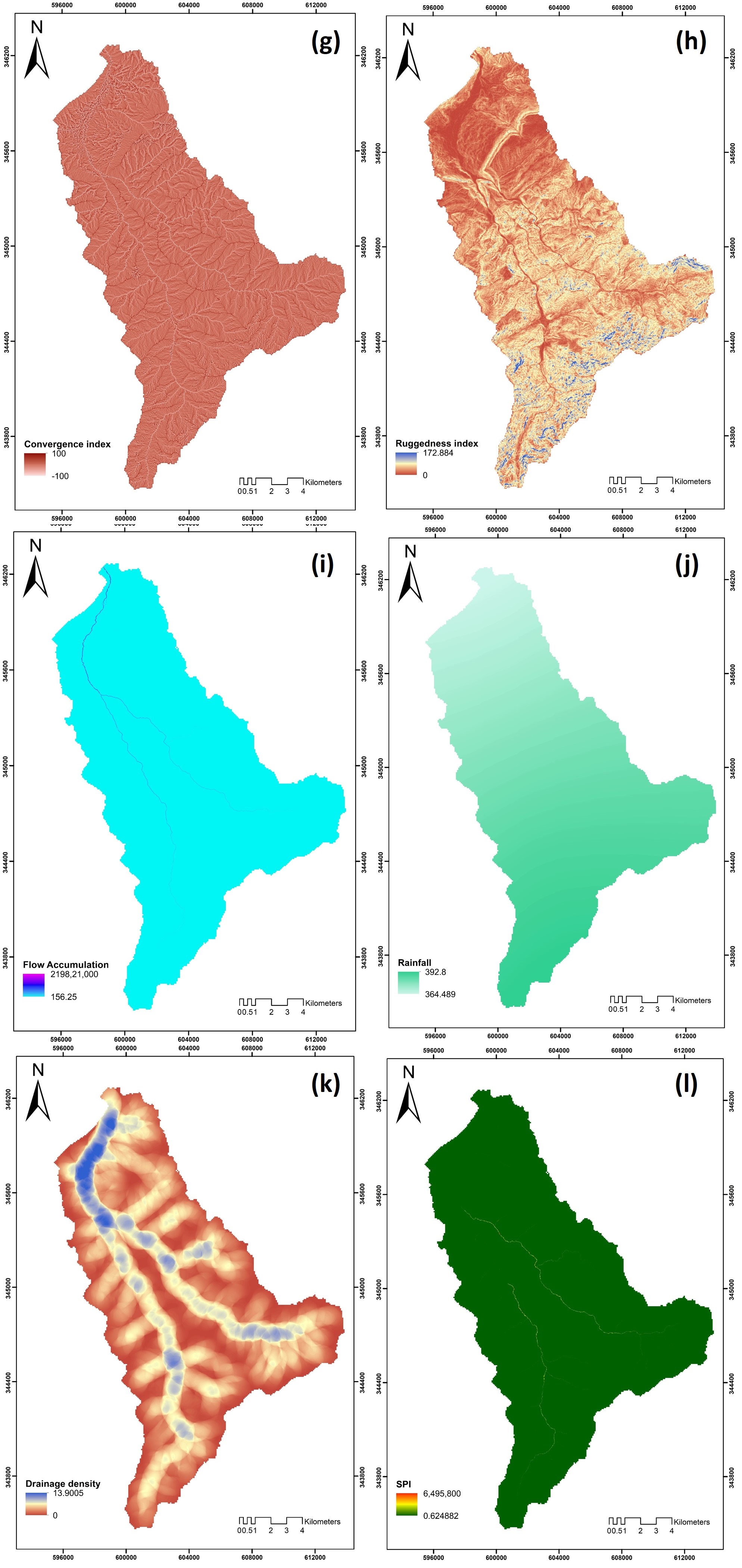}
\caption{(g) Convergence index, (h) Ruggedness index, (i) Flow accumulation, (j) Rainfall, (k) Drainge density, (l) SPI}
\label{Figure4}
\end{figure}

\begin{figure}[H]
\centering
\includegraphics[width=11cm]{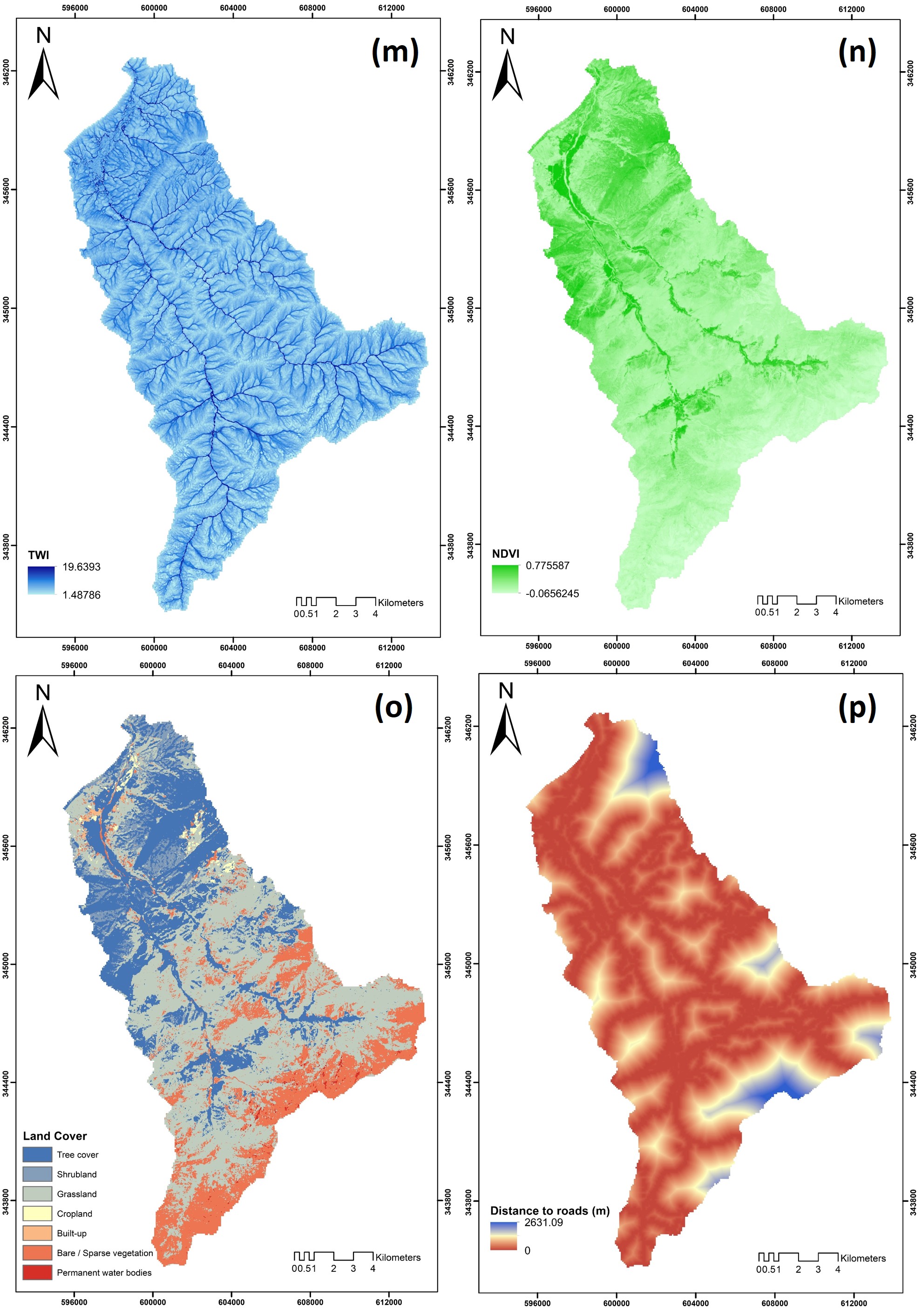}
\caption{(m) TWI, (n) NDVI, (o) Land cover, (p) Distance to roads}
\label{Figure5}
\end{figure}

The process of identifying flash flood causative factors lacks a universal framework, as it depends on the complex interactions between local environmental dynamics and historical flash flood events (\citealt{gui2023credal}). Also, these phenomena have characteristics that differ from region to region. Therefore, selecting the most sensitive flash flood conditioning factors is critical for developing an accurate flash flood-susceptibility map. In this study, 16 conditioning factors were thoughtfully selected based on a comprehensive literature review of previous studies, data availability, and the characteristics of flash flood development in the Rheraya watershed (Figure \ref{Figure3},\ref{Figure4},\ref{Figure5}). We made sure to consider different aspects of the study area, including topographical, hydrological, meteorological, environmental, and anthropogenic features (Table \ref{tab:my_label}). All factors were transformed into raster format, then aligned and resampled to 12.5 meters to ensure spatial consistency. Subsequently, data standardization was implemented to ensure a consistent scale and distribution of data values. Finally, the entire dataset was divided into training, validation, and testing sets.

\begin{table}[h]
    \caption{Description of the selected conditioning variables}
    \centering
    \small 
    \setlength{\tabcolsep}{7pt} 
    \renewcommand{\arraystretch}{0.8} 
    \begin{tabular}{lcc}
        \hline
        \textbf{Variables} & \textbf{Data source} & \textbf{Spatial resolution} \\
        \hline
        \vspace{4pt} 
        \textbf{Topographical factors} & & \\
        Elevation & \multirow{13}{*}{\raggedright Digital Elevation Model (ALOS DEM)} & 12.5 m\\
        Slope & & \multicolumn{1}{c}{12.5 m} \\
        Aspect & & \multicolumn{1}{c}{12.5 m} \\
        Curvature & & \multicolumn{1}{c}{12.5 m} \\
        Distance to river & & \multicolumn{1}{c}{12.5 m} \\
        Topographical positioning index (TPI) & & \multicolumn{1}{c}{12.5 m} \\
    Ruggedness index & & \multicolumn{1}{c}{12.5 m} \\
        \vspace{4pt} 
        \textbf{Hydrological and meteorological factors} & & \\
        Stream power index (SPI) & & \multicolumn{1}{c}{12.5 m} \\
        Topographic wetness index (TWI) & & \multicolumn{1}{c}{12.5 m} \\
        Drainage density & & \multicolumn{1}{c}{12.5 m} \\
        Convergence index & & \multicolumn{1}{c}{12.5 m} \\
        Flow accumulation & & \multicolumn{1}{c}{12.5 m} \\
        Rainfall & \multirow{1}{*}{\raggedright Hydraulic Basin Agency of Tensift (ABHT)} & \multicolumn{1}{c}{-} \\
        \vspace{4pt} 
        \textbf{Environmental and anthropogenic factors} & & \\
        NDVI & \multirow{1}{*}{\raggedright Sentinel-2 MSI - Level-2A Product} & \multicolumn{1}{c}{10 m} \\
        Landcover & \multirow{1}{*}{\raggedright ESA WorldCover 2021} & \multicolumn{1}{c}{10 m} \\
        Distance to roads & \multirow{1}{*}{\raggedright OpenStreetMap (OSM)} & \multicolumn{1}{c}{-} \\
        \hline
    \end{tabular}
    \label{tab:my_label}
\end{table}

\subsubsection{Topographical factors}\label{Methodology}

The topographical factors, including elevation, slope, aspect, curvature, distance to river, convergence index (CI), topographical positioning index (TPI), and ruggedness index (RI), were computed via processing in ArcGIS 10.2. The derived data were obtained using a 12.5 m-resolution PALSAR Phased Array type L-band Digital Elevation Model (DEM) obtained from the Alaska Satellite Facility (ASF). Elevation is a key factor in flash flood modeling, given the inverse relationship between elevation and flash floods (\citealt{fernandez2010urban}; (\citealt{bui2020verification}; (\citealt{dodangeh2020integrated}). With decreasing elevation, the terrain tends to flatten, leading to increased river water flow and vice versa (\citealt{cao2016flash}). Slope is a critical factor that affects flash floods occurrence, since it affects the speed of the water flow (\citealt{stevaux2020changing}). Regions with gentler slopes are more likely to accumulate water, potentially facilitating higher infiltration rates and reduced surface runoff speeds. These areas could be more prone to flash floods. Aspect refers to the direction of the predominant slope on the surface of the terrain. It indirectly affects flash flooding as it influences floodwater flow directions, which helps to maintain soil moisture (\citealt{chu2020ann}). Curvature delineates areas where runoff diverges or converges, impacting water flow dynamics (\citealt{ginesta2020preliminary}). Runoff speed can vary based on the slope. Convex slopes often promote faster overland flow, potentially affecting infiltration and soil saturation (\citealt{cao2016flash}), whereas concave slopes might slow down overland flow and potentially aid infiltration (\citealt{young1969soil}). Distance to river significantly influences flash flood vulnerability within a watershed as it affects the extent of flooding (\citealt{tehrany2015flood}). Areas close to the hydrographic network often face a higher susceptibility to flooding compared to those situated farther away (\citealt{butler2006supporting}; \citealt{chapi2017novel}). The topographic position index (TPI) measures the elevation difference between a specific point and its surrounding area. It helps identify topographic features like valleys, hills, or bottoms (\citealt{zwolinski2015relevance}; \citealt{newman2018evaluating}). Generally, areas with higher elevations and steeper slope, indicated by TPI, tend to be less susceptible to flash flooding compared to lower-lying regions (\citealt{alam2021flash}). The ruggedness index measures terrain roughness based on elevation variations. In flash floods, rugged terrain might affect water flow patterns, potentially influencing flash flood behavior due to altered landscape topography (\citealt{alam2021flash}).

\subsubsection{Hydrological and meteorological factors}\label{Methodology}

Hydrological and meteorological factors directly influence the occurrence and severity of flash floods. Various factors were taken into consideration in this study, such as stream power index (SPI), topographical wetness index (TWI), drainage density, convergence index, flow accumulation, and rainfall. SPI quantifies a river's erosive potential by assessing its ability to transport sediment and erode its bed during flood events, influencing the river's overall fluvial system (\citealt{knighton1999downstream}; \citealt{pamuvcar2017novel}). TWI identifies variations in spatial wetness and highlights areas susceptible to water accumulation (\citealt{chapi2017novel}; \citealt{beven1979physically}; \citealt{moore1991digital}). It helps predict potential flash flood-prone zones by considering the ratios of basin area to slope angles (\citealt{wilson2000terrain}; \citealt{nhu2020new}). Drainage density measures the total length of streams within a watershed area (\citealt{elmore2013potential}; \citealt{nguyen2020new}). Higher drainage densities suggest increased surface runoff potential, which can contribute to flooding under certain conditions (\citealt{chapi2017novel}). The convergence index measures the degree of river network convergence within a region, highlighting valleys and inter-alluvial surfaces (\citealt{costache2017flash}). Flow accumulation quantifies the cumulative flow contribution from neighboring cells towards each specific cell within a terrain. It is a crucial factor in identifying water accumulation patterns and potential flash flood-prone areas in a landscape (\citealt{abdel2020environmental}). Rainfall is one of the most critical factors that causes flash floods  
(\citealt{pourghasemi2020assessment}). These sudden floods arise when rainfall exceeds the ground's capacity to absorb water, causing rapid and unexpected flooding. In this study, we used data from 1994 to 2022 collected from two specific meteorological stations, i.e., Tahanaout and Armed, obtained from the hydraulic basin agency of Tensift (ABHT). The IDW (i.e., inverse distance weighted) interpolation method was then used to compute the precipitation variable using ArcGIS 10.2 software.

\subsubsection{Environmental and anthropogenic factors}\label{Methodology}

Environmental and anthropogenic factors included the NDVI (normalized difference vegetation index), landcover, and distance to roads. The NDVI is an index that reflects the changes in vegetation and surface water cover over time (\citealt{ahmed2017analysis}). It expresses the density of vegetation in a region and has a strong influence on flash flooding (\citealt{kumar2016flood}). Areas with lower vegetation cover often exhibit increased vulnerability to flooding (\citealt{ngo2018novel}). In this study, the NDVI was calculated using Sentinel-2 Level-2A data spanning six years, from January 1, 2017, to December 31, 2022, which included a total of 538 Sentinel-2 images. These images were filtered to include only the ones with a cloud cover of 20\% or less. Through Google Earth Engine (GEE), we generated a raster representing the mean NDVI across all these years. Land cover has a significant influence on surface runoff, infiltration, evaporation, and sediment transport (\citealt{benito2010impact};\citealt{karlsson2017natural}; \citealt{khosravi2018comparative}). It directly affects flash flood occurrence and frequency, which aids in identifying highly susceptible areas. Urban areas characterized by impermeable surfaces, for instance, are more likely to be flooded compared to areas covered by farmlands, forests, and green vegetation (\citealt{rahmati2016flood}). We used the European Space Agency (ESA) WorldCover product, which provides a global land cover map for 2021 at 10 m resolution based on Sentinel-1 and Sentinel-2 data (\citealt{zanaga2022esa}). The landcover categories included 7 classes, including tree cover, shrubland, grassland, cropland, built-up, bare or sparse vegetation, and permanent water bodies. Distance from roads can significantly impact flooding and the susceptibility of an area to flood risks. Roads often act as barriers, altering natural drainage patterns and increasing surface runoff, elevating flash flood vulnerability. Therefore, distance to roads could be an important variable of flash flood susceptibility (\citealt{nachappa2020flood}). We extracted road data from Open Street Map (OSM), and then we calculated the distance from each pixel to the road feature using the Euclidean distance in ArcGIS 10.2.

\subsection{Methodology} \label{Methodology}

In the present research, a CBAM-based CNN model was developed to assess flash flood susceptibility in the Rheraya watershed. Given their advantages, three representative deep CNNs, namely, ResNet 18 (\citealt{he2016deep}), Xception (\citealt{chollet2017xception}), and DenseNet121 (\citealt{huang2017densely}), were used as the backbone architectures (detailed in Section 2.4.2). To investigate the impact of the CBAM block on flash flood susceptibility outcomes, we integrated the CBAM at various points within these CNN architectures. Specifically, we tested three insertion points: plugging CBAM into each block of the backbone architecture (Figure \ref{Figure7}c), at the head of the backbone architecture (Figure \ref{Figure7}a), and at the tail of the backbone architecture (Figure \ref{Figure7}b).

\begin{figure}[H]
\centering
\includegraphics[width=17cm]{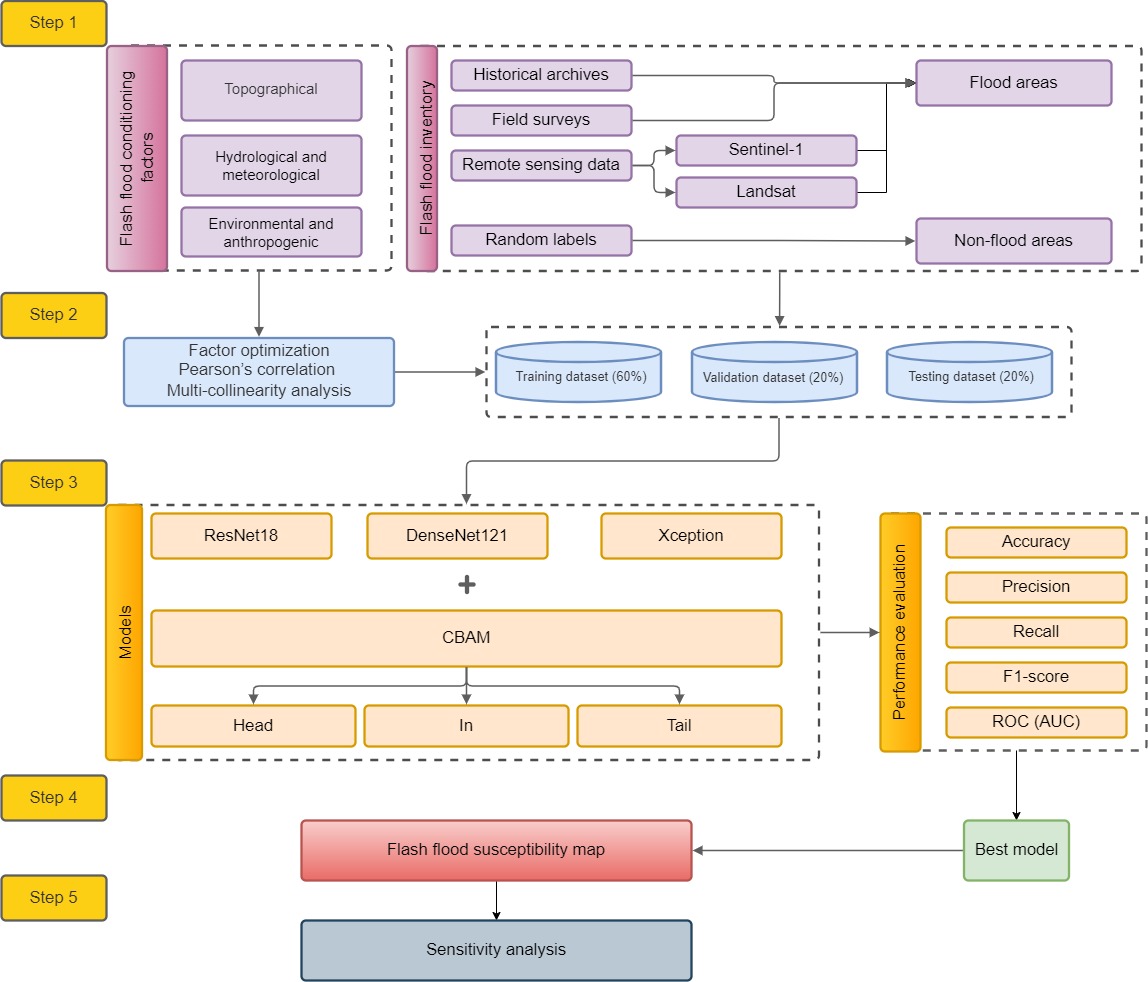}
\caption{Flowchart of the used methodology}
\label{Figure6}
\end{figure}

The workflow of the study is presented as a flowchart in Figure \ref{Figure6}, consisting of five main steps: (1) compilation of the GIS and remote sensing database, including flash flood inventory data and 16 well-selected flash flood conditioning variables; (2) identification and removal of redundant flash flood influencing variables through factor optimization (i.e., Pearson’s correlation and multi-collinearity analysis), followed by the splitting of the database into training, validation, and testing datasets; (3) developed and trained multiple CBAM-based CNNs, including the attention block at different locations within the architectures, and then evaluated and compared their performance using various statistical metrics. (4) generated a flash flood susceptibility map based on the most robust model; and (5) conducted sensitivity analysis on various flash floods with respect to the influencing variables. 

All models were trained using Keras 2.15.0, with TensorFlow 2.15 serving as the backend. Model weights and biases were initialized from scratch, and the network was trained with a batch size of 4. The initial learning rate was set to 0.001, decreasing by a factor of 10 whenever the validation set's accuracy showed no improvement for 10 consecutive epochs. Models with the lowest validation loss were chosen. Binary cross-entropy served as the loss function for the CNNs. All the analysis and model implementations were performed in Python 3.11.2.

\subsubsection{Factor optimization
} \label{Methodology}

Identifying and filtering redundant flash flood-influencing factors with limited classification ability enhances the accuracy of the process as it overcomes training challenges related to high data dimensionality (\citealt{pham2021flood}). In the context of classification or regression problems, it is crucial that independent variables exhibit minimal correlation. This increases the prediction power of the models by selecting the most significant factors and removing the irrelevant ones. Therefore, it is crucial to ensure the dependency between the flash flood conditioning variables through correlation and multicollinearity tests. In this study, we initially computed the Pearson correlation coefficient to examine correlations, followed by computing the variance inflation factor (VIF) to assess multicollinearity. These commonly used metrics have demonstrated high efficiency and accuracy in selecting relevant variables in flood susceptibility analysis (\citealt{yang2022snowmelt}; \citealt{tsangaratos2023applying}). The Pearson correlation coefficient quantifies the linear correlation between two datasets, serving as a tool to evaluate how changes in one variable align with changes in another. It has values ranging from -1 to 1, where a strong correlation is indicated by values equal to or greater than 0.7, a significant correlation falls between 0.5 and 0.7, a weak correlation is between 0.3 and 0.5, and no correlation is indicated when values are lower than 0.3 (\citealt{sedgwick2012pearson}). VIF measures multicollinearity among the independent variables, where values higher than 5 to 10 suggest a problem with multicollinearity (\citealt{kim2019multicollinearity}).

\subsubsection{Models} \label{Methodology}
\begin{enumerate}[label={(\alph*)}]
\item \textbf{ResNet18}

ResNet (Deep residual network) is one of the models developed by \citet{he2016deep} that addresses challenges encountered in the training of deep neural networks. Training deep learning models is often time-consuming and constrained by the number of layers. To overcome these issues, ResNet incorporates skip connections or shortcuts. Notably, the ResNet model exhibits superior performance compared to alternative architectures, as its effectiveness does not diminish with increasing depth. This results in more efficient computational calculations and improved network training capabilities.
The ResNet architecture employs skip connections across two to three layers, incorporating Rectified Linear Unit (ReLU) and batch normalization.\cite{he2016deep} demonstrated that ResNet excels in image classification, outperforming other models and highlighting its proficiency in extracting image features. The implementation of residual learning across multiple layers is a key aspect of ResNet's success. The residual block in ResNet is defined as: 
\begin{equation}
    y = F(x, W + x)
\end{equation}
where x is the input layer, y is the output layer, and F represents the residual map.
ResNet's residual block is effective when input and output data dimensions are identical. Additionally, each ResNet block comprises either two layers (for ResNet-18 and ResNet-34 networks) or three layers (for ResNet-50 and ResNet-101 networks). The initial two layers of the ResNet architecture resemble GoogleNet, employing a 7 × 7 convolution and 3 × 3 max-pooling with a stride of 227. In this study, we chose ResNet-18, which contains 17 convolution layers and 1 fully connected layer.

\item \textbf{DenseNet121}

Dense convolutional networks (DenseNets), introduced by \citet{huang2017densely}, represent a highly effective architecture alternative to ResNets, showcasing superior performance across various classification tasks, including those in remote sensing applications (\citealp{zhang2019full}; \citealp{tong2020channel}; \citealp{chen2021drsnet}). In contrast to ResNets, DenseNets feature dense blocks where each layer connects to every preceding layer, incorporating an additional (channel-wise) concatenated input of the previously learned feature maps. This design fosters extensive feature reuse throughout the network, resulting in well-performing and more compact models with fewer trainable parameters compared to an equivalent-sized ResNet. However, this advantage comes with the trade-off of increased memory requirements during training. Equation~\ref{eq:densenet} defines the operation of Densenet:

\begin{equation}\label{eq:densenet}
    H(x) = T_k(H_{k-1}(H_{k-2}(\ldots H_1(T_0(x)) \ldots))) 
\end{equation}

Where \(x\) is the input to the entire DenseNet; \(T_0, T_1, \ldots, T_k\) represent the transition layers; \(H_1, H_2, \ldots, H_{k-1}\) represent the dense blocks; and \(H(x)\) is the output of the DenseNet.

Among the numerous DenseNet models, in this study, DenseNet121 was used, where 121 is the total number of layers present in the architecture. DenseNet121 is a deep model comprised of four dense blocks, with the layers between consecutive blocks referred to as transition layers. These transition layers play a crucial role in altering feature-map sizes through convolution and pooling operations.

\item \textbf{Xception}

Xception, proposed by \citet{chollet2017xception}, is a convolutional neural network architecture that aims to enhance traditional convolutional layers by leveraging depthwise separable convolution layers. The term Xception, short for "Extreme Inception," reflects its inspiration from the Inception architecture. The Xception architecture has 36 convolutional layers, forming the feature extraction base of the network. Unlike standard convolutions, Xception utilizes depthwise separable convolutions, consisting of two steps: depthwise convolutions, where a separate 3x3 convolution is applied for each channel, capturing spatial dependencies; and pointwise convolutions, employing a 1x1 convolution across all channels to mix information. This design optimizes parameter usage, reducing computational costs while preserving expressive power. The architecture is modular, featuring repeated depthwise separable convolution blocks with skip connections to facilitate information flow across different depths. These skip connections aid in training very deep networks and support gradient flow during backpropagation, making Xception efficient and powerful for various deep learning tasks.

\item \textbf{Convolutional block attention module (CBAM)}

Convolutional block attention module (CBAM) is an attention module for convolutional neural networks (\citealp{woo2018cbam}). It can be integrated into any CNN to enhance its representational capacity by guiding the model to focus on relevant features and ignore non-useful background information. The architectural overview of CBAM is depicted in Figure \ref{Figure8}a. CBAM combines channel and spatial information through convolution operations, emphasizing key features in the convolution layers. This is achieved with two sequential sub-modules: the channel attention module and the spatial attention module, as shown in Figure \ref{Figure8}b. Below, we describe the details of each sub-module:

\textit{Channel attention module:} in the convolution process, each channel of intermediate features is treated as a feature detector. The channel attention module exploits inter-channel relationships to determine the importance of each channel. Initially, the spatial dimension of intermediate features undergoes average-pooling and max-pooling to generate two distinct context features. Subsequently, these features are transformed by a shared multi-layer perceptron (MLP) and combined through element-wise summation. The resulting fused feature is activated by a sigmoid function, representing the channel importance of the original feature map. Mathematically, the channel attention module's calculation process is defined as: 
\[ M_c(F) = f_{\text{sigmoid}}\left(\text{MLP}(\text{AvgPool}(F)) + \text{MLP}(\text{MaxPool}(F))\right) \ (3) \]
\textit{Spatial attention module:} in contrast to the channel attention module, the spatial attention module focuses on determining "where" in the input space significant information lies, complementing the channel attention. It aims to generate refined features by exploiting inter-spatial relations in the original features. This involves applying average-pooling and max-pooling along the channel axis, concatenating the outputs, and converting the concatenated descriptor into a spatial attention map through a 7 × 7 convolution and a sigmoid operation. The spatial attention module's computation can be expressed as: 
\[ M_s(F) = f_{\text{sigmoid}} \left( \text{Conv}([\text{AvgPool}(F); \text{MaxPool}(F)]) \right) \ (4) \]
Finally, CBAM combines attention maps derived from the channel and spatial attention modules to generate refined feature maps. The combination of the two sub-modules is described by:
\[ F' = M_c(F) \otimes F \ (5) \]
\[ F'' = M_s(F') \otimes F' \ (6) \]
where $F$ represents an intermediate feature map of the CNN, $\otimes$ denotes element-wise multiplication, and $F''$ is the resulting refined feature map.

\begin{figure}[H]
\centering
\includegraphics[width=18cm]{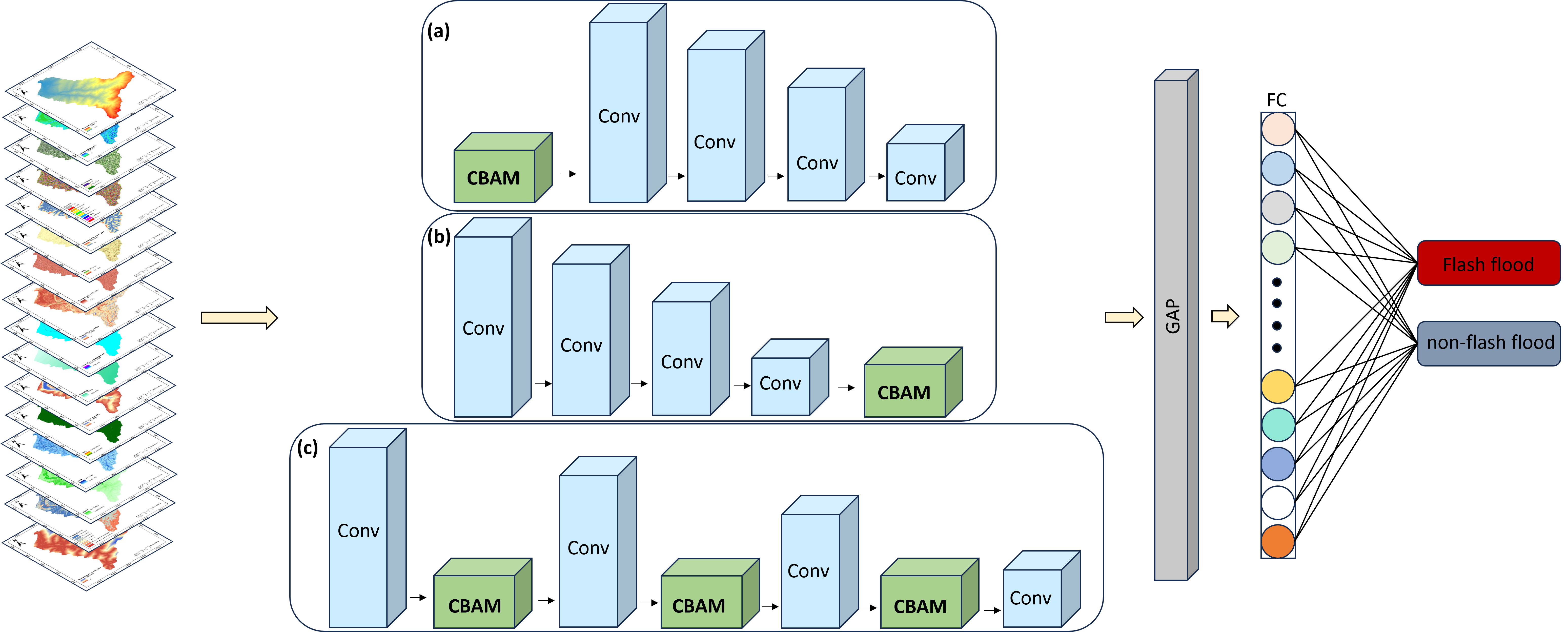}
\caption{Illustration of the CBAM-based CNN architecture, including the attention module heading the architecture (a), tailing the architecture (b), and plugged in each convolutional block (c). GAP: global average pooling, FC: fully connected layers.}
\label{Figure7}
\end{figure}

\begin{figure}[H]
\centering
\includegraphics[width=11cm]{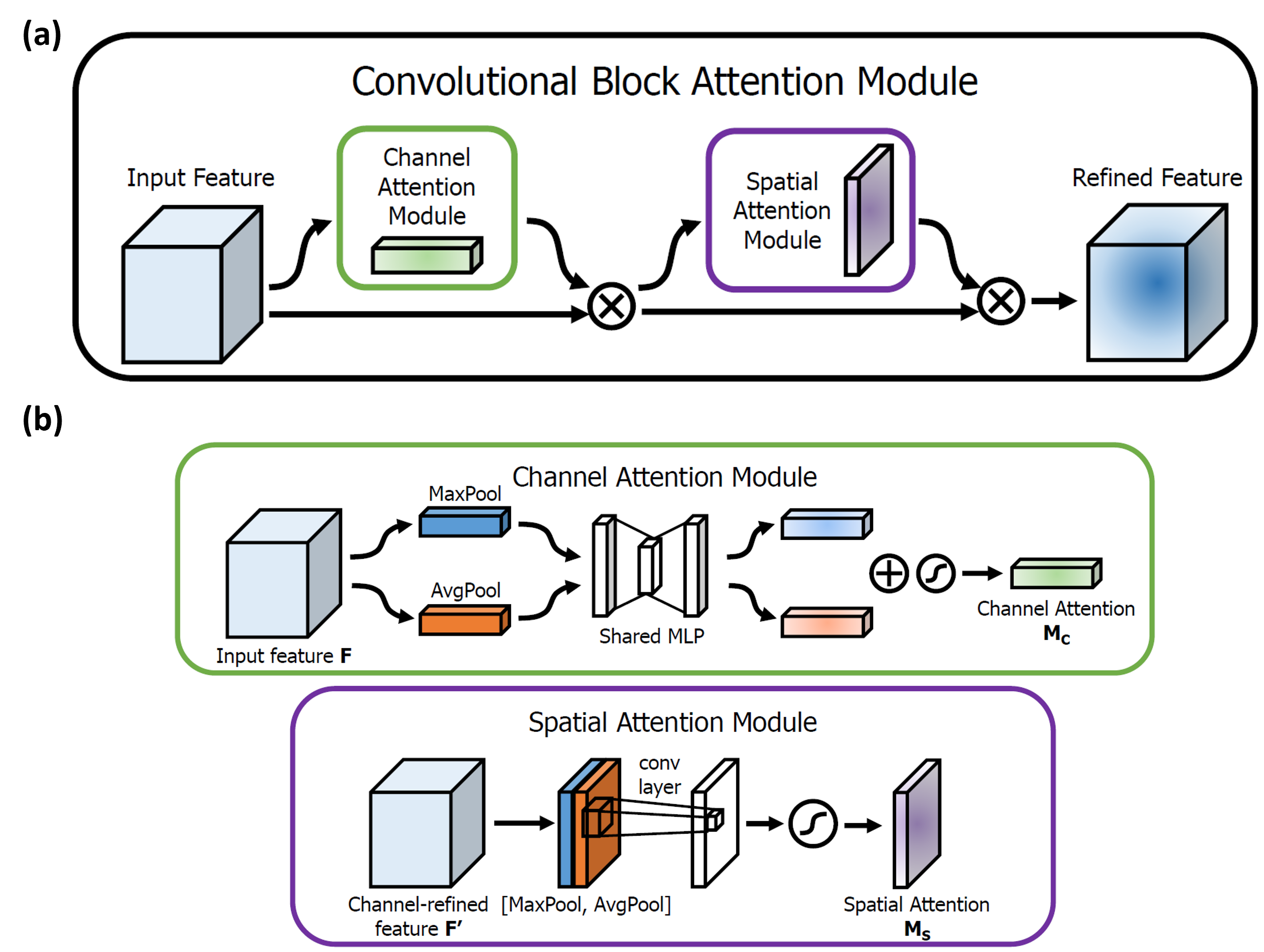}
\caption{The convolutional block
attention module architecture (CBAM)(a) and the channel attention and spatial attention sub-modules(b).}
\label{Figure8}
\end{figure}

\end{enumerate}


\subsubsection{Model validation} \label{Methodology}

To evaluate our model, four common metrics were used: accuracy (\(A\)), precision (\(P\)), recall (\(R\)), and F1-score (\(F\)). These metrics rely on values obtained from the confusion matrix, where true-positive (\(TP\)) and true-negative (\(TN\)) signify correctly predicted positive and negative samples, while false-positive (\(FP\)) and false-negative (\(FN\)) represent misclassified positive and negative samples. The calculations for these metrics are as follows:

\[ A = \frac{TP + TN}{TP + FP + TN + FN} \ (7) \]

\[ P = \frac{TP}{TP + FP} \ (8) \]

\[ R = \frac{TP}{TP + FN} \ (9) \]

\[ F = \frac{2 P R}{P + R} \ (10) \]

Additionally, the evaluation incorporated the use of the receiver operating characteristic (ROC) curve and the area under the curve (AUC). The ROC curve plots "1-specificity" on the X-axis against "sensitivity" on the Y-axis. Specifically, false positive rate (FPR) is represented by "1-specificity", and true positive rate (TPR) is represented by "sensitivity". These measures are computed as follows:

\[ FPR = \frac{TN}{TN + FP} \ (11) \]

\[ TPR = \frac{TP}{TP + FN} \ (12) \]

The AUC value ranges from 0.5 to 1. A higher AUC means better classification performance.
\subsubsection{Sensitivity analysis} \label{Methodology}
Sensitivity analysis offers a comprehensive approach to assess the impact of variables or parameters on the variability and uncertainty within numerical model outputs (\citealp{xing2021influence}). Widely adopted for statistical assessments of model characteristics, the Jackknife test holds particular significance, especially in the context of AUC-based statistical coefficients, and is recognized for its effectiveness in addressing a diverse array of practical challenges (\citealp{bandos2017jackknife}). Hence, in our study, the Jackknife test was used to evaluate the sensitivity of a specific factor to flash flooding.

In this test, we compared predictions based on all dependent variables with predictions deliberately excluding one dependent variable. Our focus was on observing the resultant reduction in AUC concerning the original value (\citealp{park2015using}). The percentage of relative decrease (PRD) emerged as a crucial metric in this analysis. The importance of the excluded factor is considered more pronounced if the PRD experiences a more substantial reduction. The PRD is determined by the formula:

\[PRD_i = \frac{100 \times |AUC_o - AUC_i|}{AUC_o}\ (13) \]

Here, \(AUC_o\) represents the original AUC value encompassing all factors, while \(AUC_i\) denotes the AUC value when the \(i\)-th factor has been excluded.

\section{Results} \label{Results}
\subsection{Factor optimization} \label{Results}

\begin{figure}[H]
\centering
\includegraphics[width=11cm]{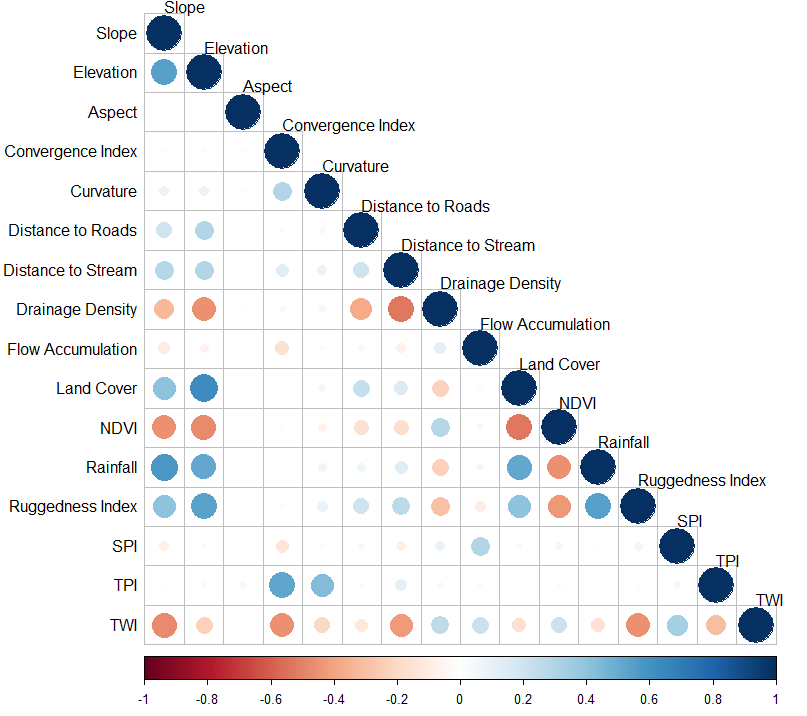}
\caption{Pearson's correlation analysis between the conditioning factors. SPI: Stream power index, TPI: Topographic positioning index, TWI: Topographic wetness index.}
\label{Figure9}
\end{figure}

The correlation between flash flood predictors was calculated using the Pearson coefficient, as shown in Figure \ref{Figure9}. The results indicate that most variables have a low correlation, suggesting they are independent. However, a slightly moderate correlation appears between landcover and elevation, slope and rainfall, NDVI and landcover, as well as distance to stream and drainage density. This correlation is not significant. The multi-collinearity analysis was conducted using the VIF to enhance the prediction power and performance of the model. The results are summarized in Table \ref{tab:2}. The VIF values range from 1.0086 to 4.7282, with rainfall exhibiting the highest value and aspect the lowest. These findings imply that there are no multi-collinearity problems among the flash flood variables. Therefore, all 16 variables were included in the present study.

\begin{table}
  \centering
  \caption{Multi-collinearity analysis (VIF) result}
  \small 
  \setlength{\tabcolsep}{7pt} 
  \renewcommand{\arraystretch}{0.8} 
  \begin{tabular}{lr}
    \toprule
    Variable & VIF \\
    \midrule
    Elevation & 4.4922 \\
    Slope & 4.5012 \\
    Aspect & 1.0086 \\
    Convergence index & 1.8449 \\
    Curvature & 1.2732 \\
    Distance to roads & 1.3563 \\
    Distance to stream & 1.6339 \\
    Drainage Density & 1.8492 \\
    Flow accumulation & 1.2999 \\
    Land cover & 2.6179 \\
    NDVI & 2.7874 \\
    Rainfall & 4.7282 \\
    Ruggedness index & 2.6248 \\
    SPI & 1.3935 \\
    TPI & 1.7260 \\
    TWI & 2.4336 \\
    \bottomrule
  \end{tabular}
  \label{tab:2}
\end{table}

\vspace*{1cm}

\subsection{Model validation} \label{Results}

The performance of the twelve models was assessed using various standard quantitative metrics (Table \ref{tab:3}). The results show that adding the CBAM attention module significantly improved performance, demonstrating its effectiveness in flash flood susceptibility modeling. In addition, the location of the CBAM block within the backbone architectures influenced accuracy, with embedding CBAM in each convolutional block yielding the best results. Among the models, DenseNet+CBAM-In achieved the highest accuracy (1 and 0.95), followed by Xception+CBAM-In, ResNet+CBAM-In, Xception+CBAM-Tail, Xception+CBAM-Head, DenseNet+CBAM-Tail, ResNet+CBAM-Head, DenseNet+CBAM-Head, ResNet+CBAM-Tail, ResNet, DenseNet, and Xception. The DenseNet+CBAM-In model also excelled in precision, recall, and F1-score metrics, indicating its ability for creating more accurate flash flood susceptibility maps.



\begin{sidewaystable}
    \centering
    \caption{The performance of CNN models with and without the CBAM attention module}
    \scriptsize 
    \setlength{\tabcolsep}{1.5pt} 
    \renewcommand{\arraystretch}{1.2} 
    \begin{adjustbox}{}
        \begin{tabular}{ccccccccccccc}
            \hline
            & \multicolumn{4}{c}{ResNet18} & \multicolumn{4}{c}{DenseNet121} & \multicolumn{4}{c}{Xception} \\ \hline
            Performance metric & Base &  +CBAM-In &  +CBAM-Head &  +CBAM-Tail & Base &  +CBAM-In & 
 +CBAM-Head &  +CBAM-Tail & Base &  +CBAM-In &  +CBAM-Head &  +CBAM-Tail \\ \hline
            Training set & \multicolumn{12}{c}{} \\ \hline
            Accuracy & 1 & 1 & 1 & 1 & 1 & 1 & 1 & 1 & 0.97 & 1 & 1 & 0.99 \\
            Precision & 1 & 1 & 1 & 1 & 1 & 1 & 1 & 0.99 & 0.98 & 1 & 1 & 0.99 \\
            Recall & 1 & 1 & 1 & 1 & 1 & 1 & 1 & 0.99 & 0.98 & 1 & 1 & 0.99 \\
            F1-score & 1 & 1 & 1 & 1 & 1 & 1 & 1 & 0.99 & 0.98 & 1 & 1 & 0.99 \\ \hline
            Testing set & \multicolumn{12}{c}{} \\ \hline
            Accuracy & 0.86 & 0.92 & 0.89 & 0.87 & 0.85 & 0.95 & 0.88 & 0.9 & 0.83 & 0.93 & 0.9 & 0.92 \\
            Precision & 0.87 & 0.92 & 0.89 & 0.88 & 0.91 & 0.95 & 0.88 & 0.9 & 0.82 & 0.93 & 0.91 & 0.92 \\
            Recall & 0.87 & 0.92 & 0.89 & 0.88 & 0.9 & 0.95 & 0.88 & 0.9 & 0.82 & 0.92 & 0.9 & 0.92 \\
            F1-score & 0.87 & 0.92 & 0.89 & 0.88 & 0.9 & 0.95 & 0.88 & 0.9 & 0.82 & 0.92 & 0.9 & 0.92 \\ \hline
            Number of parameters & 575697 & 598741 & 575844 & 584259 & 7085185 & 7524609 & 7085332 & 7217443 & 20867273 & 21604006 & 20867420 & 21393835 \\ \hline
        \end{tabular}
        \label{tab:3}
    \end{adjustbox}
\end{sidewaystable}

\begin{figure}[H]
\centering
\includegraphics[width=18 cm]{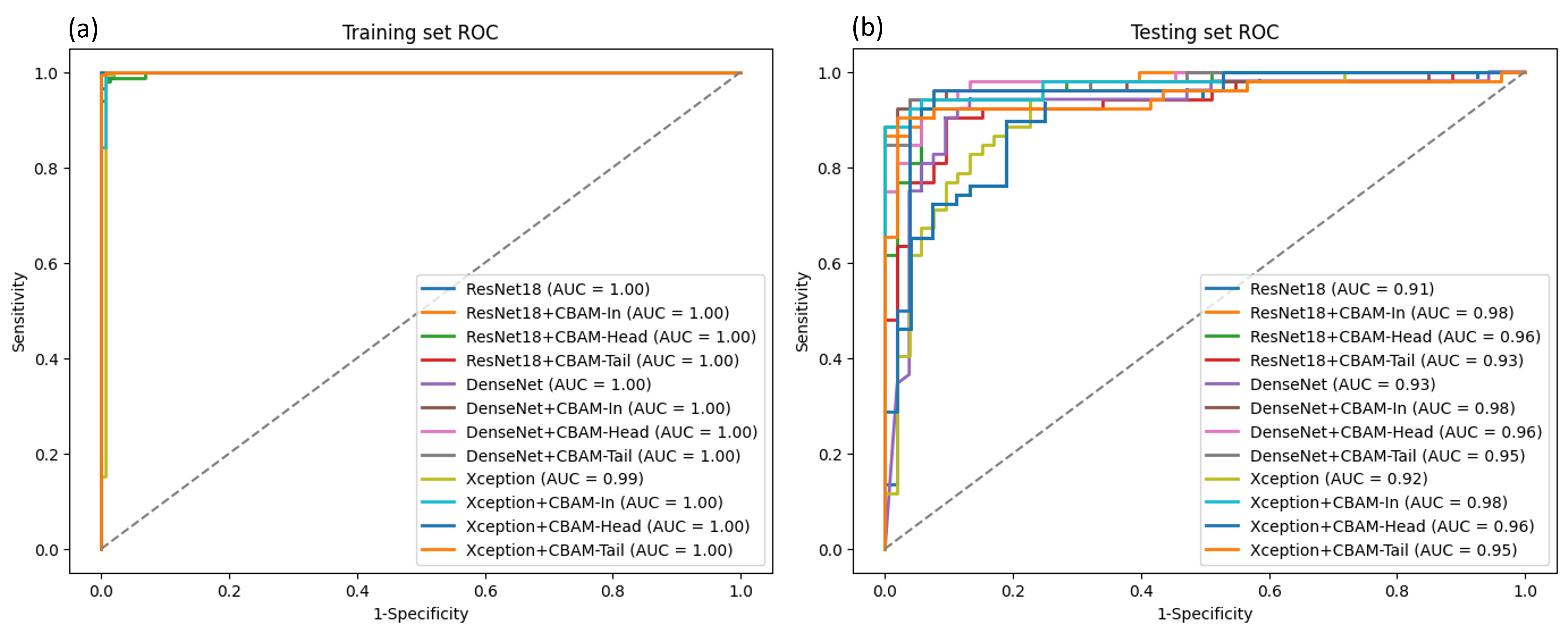}
\caption{ ROC curve and AUC value of the (a) training set and (b) testing set.}
\label{Figure10}
\end{figure}

According to the ROC results (Figure \ref{Figure10}), all models demonstrated very high performance in terms of AUC values on the training sets (AUC = 1), except for Xception, which had a slightly lower value (AUC = 0.99). On the test sets, the models varied in predictive performance, with AUC values ranging from 0.91 to 0.98. DenseNet+CBAM-In, Xception+CBAM-In, and ResNet+CBAM-In exhibited the highest AUC value of 0.98, while ResNet showed the lowest performance with an AUC of 0.91. In general, all models achieved an outstanding flash flood susceptibility assessment (AUC values surpassing 0.90) (Hosmer et al., 2013). DenseNet+CBAM-In demonstrated superior predictive and generalization capabilities compared with the other models based on all evaluation metrics. Therefore, this model was finally chosen to generate the flash flood susceptibility map of the Rheraya watershed.

\subsection{Mapping flash flood susceptibility} \label{Results}

\begin{figure}[H]
\centering
\includegraphics[width=13 cm]{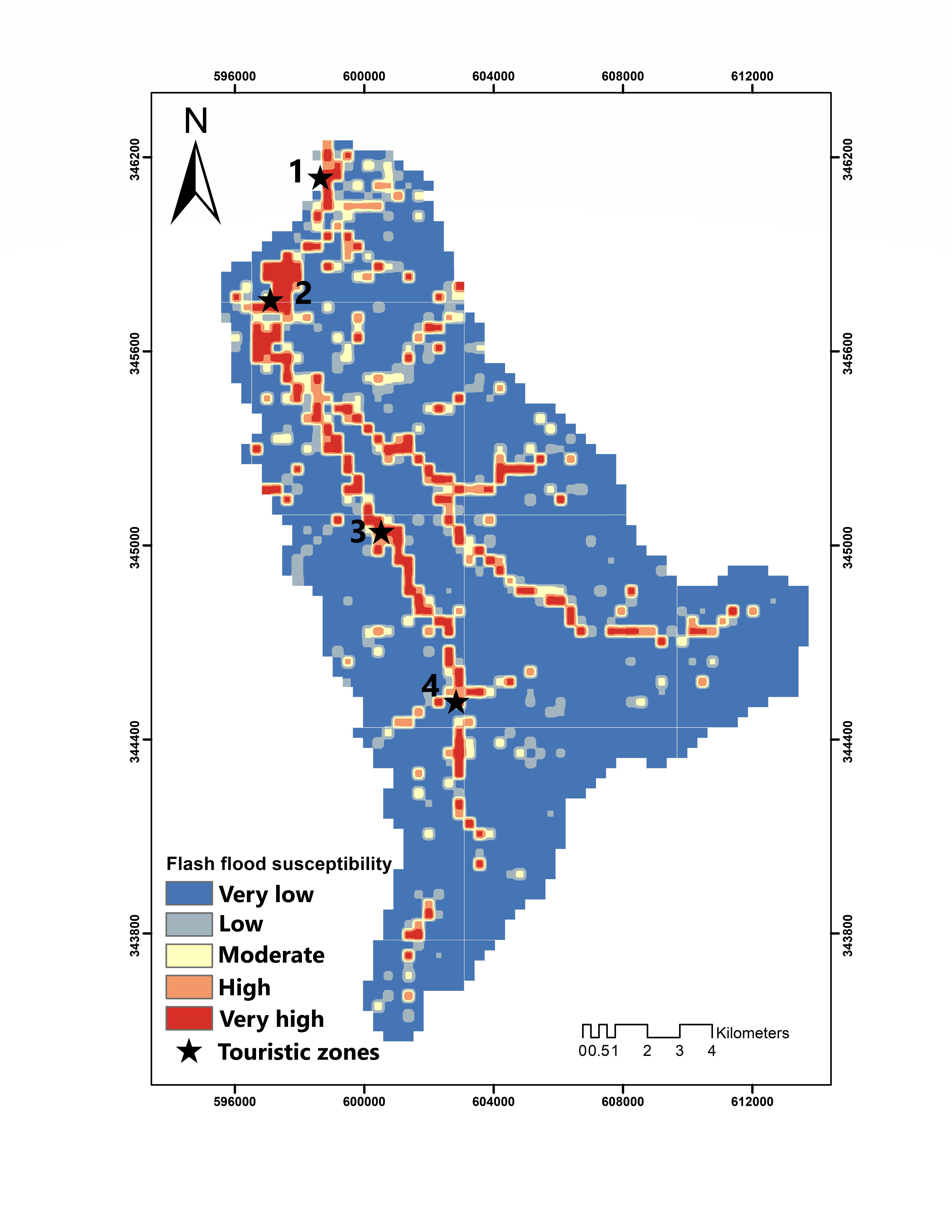}
\caption{Flash flood susceptibility map of the Rheraya watershed using the DenseNet+CBAM-In model. 1: Moulay Brahim; 2: Asni; 3: Tinine; 4: Imlil}
\label{Figure11}
\end{figure}

Flash flood susceptibility was mapped using the best-performing models (i.e., DenseNet+CBAM-In). The output was categorized into five distinct classes: very low, low, moderate, high, and very high susceptibility, using the natural break classification method (Figure \ref{Figure11}). Areas with high and very high flash flood susceptibility are primarily located along rivers, in regions with high drainage density, and concentrated in low-elevation zones. In terms of area proportion, the very high susceptibility class covered only 4.66\% of the study area, while the very low and low susceptibility classes were the most dominant, with values of 74.03\% and 10.47\%, respectively (table \ref{tab:4}). These lower susceptibility areas are characterized by high altitudes and significant distance from rivers.

The high and very high susceptibility to flash floods in the study area is notably concentrated in touristic zones such as Moulay Brahim, Asni, Imlil, and Tinitine. Therefore, these regions require significant attention from decision-makers to implement robust flood risk management strategies. To further validate our results, we assessed the distribution of flash flood events across each susceptibility zone (table \ref{tab:4}). The distribution of these flash flood events across different susceptibility classes confirms the validity of the susceptibility assessment results (\citealp{zhongping2020susceptibility}). The results show that almost 73\% of historical flash flood events were in high and very high susceptibility areas, with very few events in very low susceptibility areas. This indicates that the DenseNet+CBAM-In model effectively and reliably identified the link between historical flash flood events and flash flood-related variables, accurately pinpointing high susceptibility zones.

\begin{table}
  \centering
  \caption{Distribution of flash flood susceptibility areas and the proportion of different flash flood events across susceptibility classes.}
  \small 
  \setlength{\tabcolsep}{7pt} 
  \renewcommand{\arraystretch}{0.8} 
  \begin{tabular}{ccc}
    \toprule
    Class & \makecell{Percentage of flash flood \\ susceptible areas (\%)} & \makecell{Proportion of different \\ flash flood events (\%)}\\
    \midrule
    Very low & 74.03 & 3.15\\
    Low & 10.47 & 4.33 \\
    Moderate & 6.28 & 18.90\\
    High & 4.56 & 20.87\\
    Very high & 4.66 & 52.76 \\
    \bottomrule
  \end{tabular}
  \label{tab:4}
\end{table}

\subsection{Sensitivity analysis} \label{Results}

\begin{figure}[H]
\centering
\includegraphics[width=18 cm]{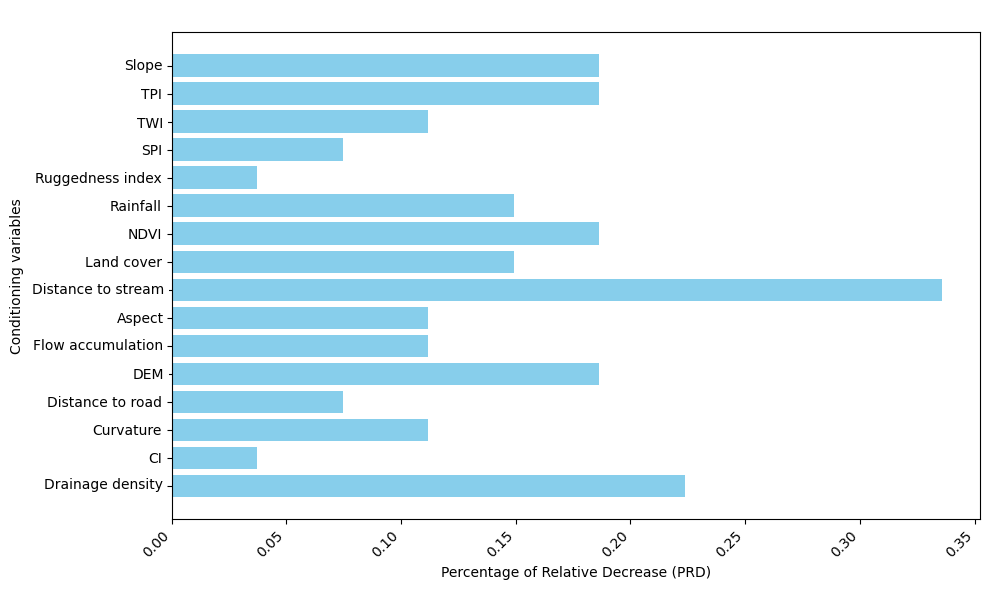}
\caption{Sensitivity analysis results. TPI: Topographic position index; TWI: Topographic wetness index; SPI: Stream power index; CI: Convergence index.}
\label{Figure12}
\end{figure}

The Jackknife test method was used to analyze the sensitivity of 16 factors related to flash flooding. The findings indicate that the distance to streams and drainage density have the most significant impact on flash floods in the Rheraya watershed, showing very high sensitivity (PRD> 0.20) compared to other factors (Figure \ref{Figure12}). Elevation, slope, NDVI, and TPI also demonstrate high sensitivity, with PRD values around 0.18. Other factors such as TWI, SPI, rainfall, landcover, aspect, flow accumulation, distance to roads, and curvature demonstrate a moderate influence on flash floods (0.07 <PRD> 0.12). In contrast, the ruggedness index and convergence index show the lowest sensitivity values (PRD= 0.04), suggesting they have little contribution to flash flood susceptibility in the study area. 

\section{Discussion} \label{Discussion}

To effectively prevent and mitigate flood hazards, it is essential to adopt suitable flood modeling techniques. Thus, exploring new and robust methods for assessing flash flood susceptibility is crucial. In this study, we present an attention-based deep learning model to assess flash flood susceptibility in the Rheraya watershed, a flood-prone region. We compared its performance with state-of-the-art baseline CNN architectures, such as ResNet18, DenseNet121, and Xception. We showed that integrating the CBAM block into the CNN architecture significantly improves model performance. In addition, we demonstrated that the location of the CBAM attention module within the architecture significantly influences the performance of the flash flood susceptibility models. To the best of our knowledge, this is the first study to apply attention-based deep learning to flash flood susceptibility and assess the impact of the attention-block module's location within the architecture on flash flood susceptibility model performance.

In recent years, artificial intelligence has advanced significantly in flood susceptibility modeling, with traditional machine learning methods like RF, KNN, and SVM being effectively employed to identify highly flood-prone areas (\citealp{abu2022random}; \citealp{el2021machine}; \citealp{tehrany2015flood}). However, deep learning techniques, particularly CNNs, have shown superior performance in automatically extracting high-throughput features for flood susceptibility mapping (\citealp{bui2020novel}; \citealp{costache2020novel}; \citealp{shahabi2021flash}; \citealp{tsangaratos2023applying}; \citealp{yang2022snowmelt}). Despite their robustness, CNNs still suffer from drawbacks, such as gradient explosion or vanishing in deeper networks and overfitting in wider ones (\citealt{joshi2019issues}). To mitigate these challenges, our study incorporated an attention module, specifically CBAM, into various deep learning networks, aiming to enhance effective feature weighting and reduce invalid weight features. Our results indicate that attention-based CNNs consistently outperform CNNs without an attention mechanism. This was also shown by \citet{mao2023attention} and \citet{ravi2023attention}, where authors showed that attention-based CNNs outperformed non-attention-based CNNs in distinguishing benign from malignant breast lesions and in malware classification, respectively. Similarly, \citet{alirezazadeh2023improving} compared the performance of multiple well-known CNN architectures, such as EfficientNetB0, MobileNetV2, ResNet50, InceptionV3, and VGG19, with and without the convolutional block attention module (CBAM) for plant disease classification and showed that EfficientNetB0+CBAM has outperformed all the other models. In addition, we evaluated the impact of CBAM placement within CNN architectures, comparing its integration in each convolutional block, in the head, and in the tail of the CNN architectures. The findings demonstrate that models integrating the CBAM block in each convolutional block yielded the highest accuracy and AUC values compared to those with CBAM only tailing or heading the architecture. For the best performing model, DenseNet121 integrated with CBAM in each convolutional block outperforms all the other models in flash flood susceptibility modeling in terms of accuracy, precision, recall, F1-score, and AUC values. It achieved an accuracy of 0.95 and an AUC value of 0.98, followed by Xception+CBAM-In (accuracy= 0.93; AUC= 0.98), ResNet+CBAM-In (accuracy= 0.92; AUC= 0.98), Xception+CBAM-Tail (accuracy= 0.92; AUC= 0.95), Xception+CBAM-Head (accuracy= 0.9; AUC= 0.96), DenseNet+CBAM-Tail (accuracy=0.9; AUC= 0.96), ResNet+CBAM-Head (accuracy= 0.89; AUC= 0.96), DenseNet+CBAM-Head (accuracy= 0.88; AUC= 0.95), and ResNet+CBAM-Tail (accuracy= 0.87; AUC= 0.93). In summary, our study demonstrated that integrating the CBAM attention module into each convolutional block significantly enhances the accuracy and reliability of CNNs for flash flood susceptibility modeling, highlighting the module's broad applicability and effectiveness across various architectures.

The sensitivity analysis identified several key factors affecting flash flood susceptibility in the Rheraya watershed, including elevation, slope, distance to streams, NDVI, TPI, rainfall, and drainage density. This underscores the intricate nature of flash floods, which arise from the interplay of topographical features, land use characteristics, and hydrological conditions. Determining the most significant contributor to flash floods in our area is crucial for implementing focused and efficient mitigation strategies, allocating resources wisely, and accurately assessing risks. The distance to streams was found to be the most influential factor in this context. 
This is consistent with the results of \cite{elghouat2024integrated}, where the authors highlighted that the distance to river and the drainage density were the most influencing factors in flash floods in the Rheraya watershed. Numerous studies have also validated the significant role of these factors in flood occurrence (\citealp{rahmati2016flood}; \citealp{pham2020gis};\citealp{bansal2022evaluating}; \citealp{chaulagain2023flood}). This is because areas close to riverbanks are more vulnerable due to their higher exposure to swift water flow and their function as natural drainage paths. Concerning the resulting flash flood susceptibility map, it indicates that areas at low elevations, with high drainage density, and situated along the river network are particularly prone to flash floods. The Rheraya watershed, known for its scenic landscapes and popular tourist destinations (\citealt{elfels2018flood}), shows a significant portion of these tourist areas, such as Moulay Brahim, Asni, Tinitine, and Imlil, as highly vulnerable to flash floods. These regions urgently require enhanced disaster preparedness and mitigation efforts. Strengthening disaster management infrastructure in these locations is crucial, including the installation of additional meteorological stations to improve rainfall measurement and flood forecasting capabilities. In addition, urban development regulations must be enforced to ensure that new construction is flood-resistant and not located in high-risk zones, particularly near riverbanks.

While this research highlighted the efficacy of attention-based CNNs for pinpointing high-susceptible flash flood zones in the ungauged Rheraya watershed, it has certain limitations. Our analysis was confined to attention-based CNNs and baseline CNN architectures, which restricted the generalizability of our results. In future work, other deep learning models such as transformers and graph neural networks, in addition to the common machine learning algorithms, should also be included in the comparison to further assess and validate our approach. The selection of flash flood factors is also crucial in susceptibility assessment. Using an optimal number of conditioning variables is vital, as an excessive number of features can significantly impact the model's performance (\citealt{pham2021flood}). Future work could consider leveraging additional data sources and validating models under varied conditions to improve flash flood susceptibility mapping accuracy. Climate models, for instance, could provide data for different future scenarios, which could enhance flash flood susceptibility maps. Furthermore, collaborative data collection and sharing efforts could improve data availability, ranging from local to global flood inventories, enabling model performance evaluations across diverse spatial scales and testing their adaptability and generalizability. The severity of flash floods is greatly influenced by rainfall characteristics (\citealt{saharia2021impact}). Thus, future studies should also consider incorporating rainfall-related factors like intensity, duration, spatial distribution, and temporal patterns into the models. Moreover, the frequency of floods is a crucial factor for flash flood susceptibility (\citealt{zhou2019assessing}), yet it appears to be underrepresented in flood-related machine learning and deep learning approaches. Finally, our study's reliance on data from a single site limits the generalizability of our findings to other ungauged basins. To ensure the scalability and robustness of our approach, it is imperative to evaluate the accuracy of our method across basins with different geo-environmental settings.

\section{Conclusion} \label{Conclusion}
Developing accurate flash flood susceptibility maps with advanced modeling approaches is crucial for protecting communities, planning infrastructure, and mitigating risks. In this study, we investigated the convolutional block attention module (CBAM) to enhance the performance of CNN networks for flash flood susceptibility modeling by integrating it into baseline architectures such as ResNet18, DenseNet121, and Xception and assessed their performance with and without the attention module. We also evaluated the influence of the attention module’s location within the architecture by integrating it at the head, tail, and within each convolutional block of the backbone architecture. Sixteen flash flood conditioning factors and 522 flash flood inventory points were used as inputs for the modeling. We demonstrated that the attention-based approach outperformed non-attention-based CNNs in flash flood susceptibility modeling, with the best results achieved by integrating the attention module within each convolutional block of the backbone architecture. The best-performing model, the attention-based DenseNet121, achieved an accuracy of 0.95 and an AUC value of 0.98 on the test set, indicating the robustness and generalizability of our approach for accurately identifying the relationship between conditioning factors and flash flood occurrence. Sensitivity analysis revealed that distance to river and drainage density are the most influential factors in flash flood susceptibility. Our maps indicate that highly touristic zones such as Moulay Brahim, Asni, Imlil, and Tinine are highly susceptible to flash floods, consistent with previous studies in the Rheraya watershed. Our analysis was limited to attention-based CNNs and baseline architectures, future work should include other deep learning models like transformers and graph neural networks, as well as traditional machine learning algorithms, to further validate the generalizability of our approach. Additionally, leveraging more data sources and validating models under varied conditions can improve mapping accuracy. Overall, our approach can provide valuable support for policymakers in planning and managing flash flood risks in ungauged basins with high-traffic areas. 

\section*{Declaration of Competing Interest}
The authors declare that they have no known competing financial interests or personal relationships that
could have appeared to influence the work reported in this paper.

\section*{Data availability}
The dataset generated during and/or analyzed during the current study is available from the corresponding author on reasonable request.

\section*{Code availability}
The scripts used to perform this study will be made avaiable to the public after the acceptation of the
paper.

\bibliography{References}

\end{document}